\documentclass[conference]{IEEEtran}
\IEEEoverridecommandlockouts
\usepackage{cite}
\usepackage[numbers]{natbib}
\usepackage{amsmath,amssymb,amsfonts}
\usepackage{algorithmic}
\usepackage[pdftex]{graphicx}
\usepackage{textcomp}
\usepackage{xcolor}
\def\BibTeX{{\rm B\kern-.05em{\sc i\kern-.025em b}\kern-.08em
    T\kern-.1667em\lower.7ex\hbox{E}\kern-.125emX}}

\begin{document}

\title{Denoising Diffusion as a New Framework for Underwater Images}

\author{\IEEEauthorblockN{Nilesh Jain}
\IEEEauthorblockA{\textit{Computer Science and Applied Mathematics} \\
\textit{University of Witwatersrand}\\
Johannesburg, SA \\
nilesharnaiya@gmail.com}
\and
\IEEEauthorblockN{ Elie Alhajjar}
\IEEEauthorblockA{\textit{Engineering and Applied Sciences} \\
\textit{RAND Corporation}\\
Arlington, VA, USA \\
eliealhajjar@gmail.com}
}
\maketitle
\begin{abstract}
\looseness=-1
Underwater images play a crucial role in ocean research and marine environmental monitoring since they provide quality information about the ecosystem. However, 
the complex and remote nature of the environment results in poor image quality with issues such as low visibility, blurry textures, color distortion, and noise. In recent years, research in image enhancement has proven to be effective but also presents its own limitations, like poor generalization and heavy reliance on clean datasets. 
One of the challenges herein is the lack of diversity and the low quality of images included in these datasets. Also, most existing datasets consist only of monocular images, a fact that limits the representation of different lighting conditions and angles. In this paper, we propose a new plan of action to overcome these limitations. On one hand, we call for expanding the datasets using a denoising diffusion model to include a variety of image types such as stereo, wide-angled, macro, and close-up images. On the other hand, we recommend enhancing the images using Controlnet to evaluate and increase the quality of the corresponding datasets, and hence improve the study of the marine ecosystem.
\end{abstract}


\section{Problem and Motivation}
\looseness=-1
\begin{figure*}[t!]
    \centering
    \includegraphics[width=0.8\linewidth]{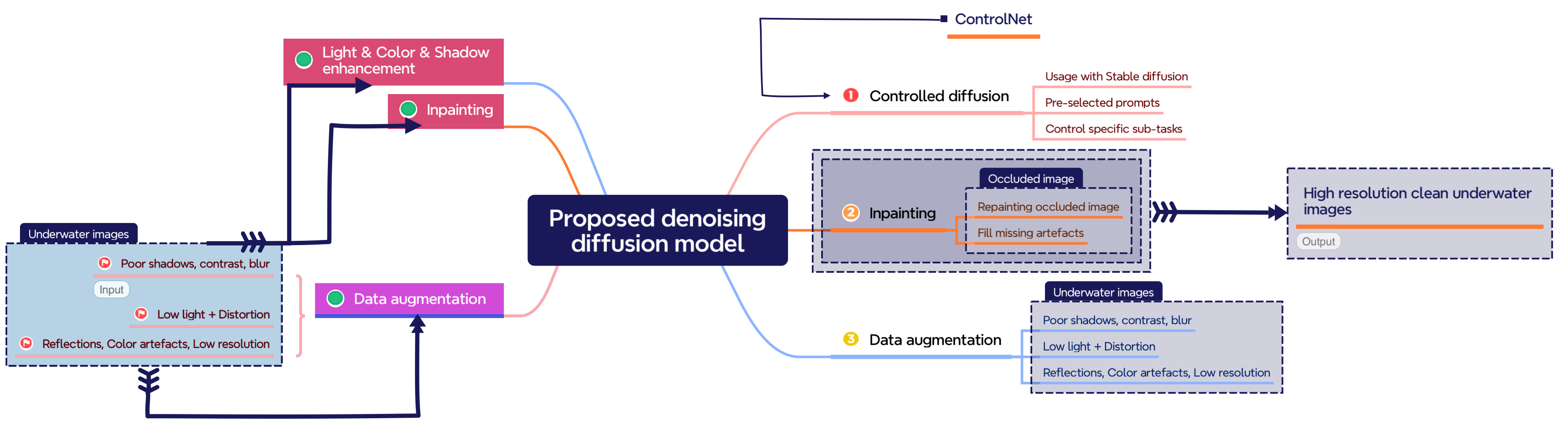}
    \caption{Denoising Diffusion Pipeline}
    \label{fig:proposedchart}
\end{figure*}
Underwater imagery of marine life often suffers from various challenges, including color distortion, background and light noise, contrast issues, blurriness, occluded objects, and suboptimal lighting conditions \cite{real_world_problems,umgan}. High-quality underwater image data is crucial for understanding and conserving marine ecosystems – a vital component in climate regulation and ocean protection. However, the current methods employed in capturing quality images are very likely to yield low quality results due to the nature of the terrain. Obtaining and processing real-world underwater datasets is a challenging process that requires extensive resources, including human labor and high computing power for research. Also, various factors such as the type of image enhancement algorithms used, location specifics, and environmental conditions significantly affect the quality of these images and hinder our adaptation to the climate emergency on a larger scale.
When collecting images in underwater environments, it is natural to deal with issues like low light conditions and high levels of background noise. While different steps are being implemented to improve visibility by using illumination devices, these measures turned out to be insufficient. Moreover, during training of deep learning models on image datasets, issues like shadows, contrast variations, and inadequate lighting impact the quality of the data even more, and hence affect the output of the model itself \cite{data_affects_model}.
Underwater datasets like WaterGAN \cite{Li_2017,underwater_color_using_haze} are commonly used for studying marine life and identifying potential threats to biodiversity. Multiple solutions such as image enhancement, color correction, and restoration have been implemented \cite{deepimage_ill, underwater_color_using_haze}, but they often fall short of effectively addressing the issue as a whole \cite{correlation_image}.
\section{Related Work/Datasets}
\looseness=-1
Previous work has explored various deep learning techniques to grapple with the challenges associated with enhancing low light underwater images. \citet{deep_traditional} developed a hybrid model that combines deep learning and traditional statistical analysis to correct color distortions in red and green channels, using an attention mechanism for selecting specific image parts for enhancement. Their approach relies heavily on training with synthetically distorted images using CycleGAN \citep{cyclegan}, which limits its generalization performance. Some studies have focused on generating synthetic degraded images from a model-based simulation that can generate images with a range of parameters and simulates the processes of absorption and scattering which is common in underwater environments 
\cite{simulation_underwater}. 
More recently, researchers have turned to generative adversarial networks \cite{og_gan} for image enhancement tasks. \citet{Li_2017} focused on real-time color correction of underwater monocular images using an unsupervised two-staged network architecture limited by the assumption of a centered vignetting pattern within the dataset. 
These deep learning approaches are thriving in significantly improving image quality in underwater scenes due to large training size availability. While GANs have shown promise, traditional haze removal methods \cite{underwater_color_using_haze, jimaging5100079} attempt to increase the visibility of the scene and correct the color shift that is inherently present in hazy images but is unreliable for stereo or wide-angled images. Meanwhile, convolutional neural networks have been used for identifying sea cucumber from video data \cite{sea_cucumber}, but the main challenge is to increase the training dataset to train the CNN, which is data-hungry and requires a clean enhanced dataset for achieving accurate results. Synthetic images have been developed as a data augmentation method to enrich the underwater image sets and help increase the training set reliability and generalizability of models -- a major issue for GANs \cite{trabucco2023effective}. 
Aside from addressing synthetic data creation, denoising diffusion models have demonstrated their usefulness in improving lighting conditions with low light images. The authors in \cite{panagiotou2023denoising}  uncovered a technique inherent within probabilistic diffusion models that assists in illuminating darker portions of an image using a noise detector given low light images, leading to good visibility levels in images after denoising. \citet{review_on_dl} demonstrated the effectiveness of deep learning-based image restoration in improving the detection accuracy of specific objects in underwater scenes.

\section{Proposed Methodology} 
\looseness=-1
We propose a novel multi-faceted denoising diffusion pipeline as seen in Figure 1 using Stable Diffusion v2.0 \cite{stable_diffusion} -- a latent diffusion model used to generate artwork from given text or prompts which is similar to Dall-e \cite{original_dalle2} in conjunction with Controlnet \cite{controlnet}. Our method consists of three sub-parts: image enhancement and artifact removal, inpainting, and data augmentation.
To address various image artifact issues such as noise, lighting artifacts, color contrast, haze, color distortions, and sharpness correction in underwater images, we utilize Controlnet \cite{controlnet} for better conditioning and control of the overall image preprocessing process. Controlnet is a variant of diffusion models that can control the output of the model and use depth maps with stable diffusion to enhance light and removal of objects in images through careful prompt engineering. Combining these techniques allows for targeted editing of specific areas within an image based on user prompts. The generation will be guided by prompts which will be pre-defined for each sub-task of image enhancement.
This combination also enables the removal of unwanted objects or artifacts, as well as the enhancement of important features or details. 

\subsection{Image enhancement and artifacts removal}
In \cite{panagiotou2023denoising}, the authors use a trick inherent in denoising diffusion models where it can illuminate the dark parts of a low-light image in post-processing.
The combination of \cite{panagiotou2023denoising} with Controlnet promises to eradicate some of the artifact problems. Sample prompt engineering will help remove shadows, light reflections, contrast, etc.
Specific prompts will help remove specific background objects and noise from the input image using Controlnet, resulting in a more accurate portrayal of the images since clear data is essential to perform any significant analysis. \citet{pearl2023svnr} show promising results by using a noisy input image as the starting point for denoising diffusion models to remove non-Gaussian noises from the image. We envision Controlnet with noisy maps can achieve similar outputs of denoising.  
\subsection{Inpainting }
One application area that diffusion methods have popularized is editing parts of an image and repainting existing real images with prompting. Inpainting \cite{pondavenconvolutional} with diffusion is one such approach that allows the user to mask specific parts to edit to fill in missing information or repaint damaged parts \cite{palette}. This enables the editing of occluded objects and artifacts, as well as the improvement of existing images under constraints. Combining Controlnet with inpainting could have improved outcomes to create clean accurate images.  

\subsection{Data augmentation}

Existing approaches to augment data using GANs to generate synthetic images from scratch have become quite popular \cite{fast_gan, umgan}. A recent paper \cite{trabucco2023effective} used real images instead of synthetic images in the generation process of diffusion models, rather than starting from scratch. We propose that with a few parameter changes, Controlnet diffusion can help generate diverse samples in different lighting conditions, angles, etc. to train robust deep learning models.

\section{Impact and Conclusion}
Our research has a far-reaching impact in marine archaeology as it helps in underwater tracking and exploration of species to understand migratory patterns due to climate-related disasters. 
In addition to that, it allows scientists to explore deeper into marine life in the ocean, where the identification of artifacts relies solely on the visibility of images.
The proposed methodology also aids the investigation into available marine resources, counting species, monitoring sea organisms, conducting geological surveys, and improving object recognition capabilities. The generalizability of our models will help researchers craft and contribute robust deep learning models for advancements in marine engineering since it can deal with monocular, stereo, wide-angled, macro and close-up images that will significantly boost the quality of the datasets for processing and analysis.

\bibliographystyle{rusnat}
\bibliography{main.bib}

\end{document}